\documentclass{svproc}

\usepackage{url}
\usepackage{graphicx, subfig}
\graphicspath{{./img/}}

\begin{document}
	\mainmatter              
	\title{Intent Classification in Question-Answering Using LSTM Architectures}
	\titlerunning{Intent Classification in Question-Answering Using LSTM Architectures}  
	\author{	Giovanni Di Gennaro\inst{1}, Amedeo Buonanno\inst{2}, Antonio Di Girolamo\inst{1}, \\ 
			Armando Ospedale\inst{1}, Francesco A.N. Palmieri\inst{1}}
	\authorrunning{Di Gennaro, Buonanno, Di Girolamo, Ospedale, Palmieri} 
	\tocauthor{Giovanni Di Gennaro, Amedeo Buonanno, Antonio Di Girolamo, Armando Ospedale, Francesco A.N. Palmieri}
	\institute{	Universit\'a degli Studi della Campania ``Luigi Vanvitelli'', \\ Dipartimento di Ingegneria \\
	              	via Roma 29, Aversa (CE), Italy\\
	              	\and
	              	ENEA, Energy Technologies Department - Portici Research Centre,\\
		 	P. E. Fermi, 1, Portici (NA), Italy;\\
	\email{ 	\{giovanni.digennaro, francesco.palmieri\}@unicampania.it \\
			\{amedeo.buonanno\}@enea.it\\
		     	\{antonio.digirolamo, armando.ospedale\}@studenti.unicampania.it 	} 
	}
	\maketitle   

	\begin{abstract}
		Question-answering (QA) is certainly the best known and probably also one of the most complex problem within 
		Natural Language Processing (NLP) and artificial intelligence (AI). Since the complete solution to the problem of 
		finding a generic answer still seems far away, the wisest thing to do is to break down the problem by solving  
		single simpler parts. Assuming a modular approach to the problem, we confine our 	research to intent classification 
		for an answer, given a question. Through the use of an LSTM network, we show how this type of classification can 
		be approached effectively and efficiently, and how it can be properly used within a basic prototype responder.

	\end{abstract}

	\section{Introduction}
		Despite the remarkable results obtained in the different areas of Natural Language Processing, the solution to the 
		Question-Answering problem, in its general sense, still seems far away \cite{Jurafsky2019}. This lies in the fact 
		that the search for an answer to a specific question requires many different phases, each of which representative 
		of a separate problem. For this reason, in this work, we have approached the problem confining our attention to the 
		classification of the intent of a response, given a specific question. In other words, our objective is not to classify the 
		incoming questions according to their meaning, but rather referring them to the type of response they may require.

		The interest in this case study is not limited to the achievement of the aforementioned objective, but also to the 
		building of a processing block that could be inserted in a larger architecture of an autonomous 	responder.
		Furthermore, current chatbot-based dialogue interfaces can also already take advantage of these type of
		structures \cite{Shen2016}, \cite{Meng2017}. 

		By acting just as an aid in the separation of intents, it is possible to assess incoming questions in a more 
		targeted manner. The objective is to conceive  simple systems, such as those based on AIML (Artificial Intelligence 
		Markup Language), to circumvent the extremely complex problem of evaluating all possible variations of the input.
 
		In the following, after an introduction to the LSTM and its main features, the predictive model is presented in 
		two successive steps. 	Finally, we report an example of a network trained to respond after it has been trained on 
		previous dialogues.

	\section{A particular RNN: The LSTM}
	\label{sec:RNN-Intro}

		The understanding of natural language is directly linked to human thought, and as such is persistent. During reading, 
		the comprehension of a text does not happen simply through single word understanding, but mostly through the way 
		in which they are arranged. In other words, there is the need to model the dynamics with which individual words arise.

		Unfortunately, traditional feedforward neural networks are not directly fit to extrapolate information from the 
		temporal order in which the inputs occur, as they are limited to considering only the current input for their 
		processing. The idea of considering blocks of input words as independent from each other is too limited within the NLP 
		context. Indeed, just as the process of reading for a human being does not lead to a memorization of all the words 
		of the text, but to the extraction of the fundamental concepts expressed, we need a ``memory" that is not limited 
		only to explicit consideration of  the previous inputs, but compresses all the relevant information acquired at each 
		step into state variables.

		This has naturally led to the so-called Recurrent Neural Networks (RNN) \cite{RNN} that, by introducing the concept 
		of ``internal state" (obtained on the basis of the previous entries), have already shown great promise, especially in 
		NLP tasks. The RNNs contain loops inside them, which allow the previous information to pass through the various 
		steps of the analysis. An explicit view of the recurrent system can be obtained by unrolling the network activations 
		considering the individual subnets as copies (updated at each step) of the same network, linked together to form a 
		chain in which each one sends a message to his successor. 

		In theory, RNNs should be able to retain in their states  the information that the learning algorithm finds in a 
		training sequence. Unfortunately, in the propagation of the gradient backwards through time, the effect of desired 
		output values that contribute to the cost function can become so small that, after only a few steps, there is no
		longer sufficient contribution to parameter learning (vanishing gradient problem). This problem, explored by 
		\cite{Bengio94}, means that the RNNs can only retain a short-term memory, because the parts of the sequence 
		further away in time is gradually less important. This makes the RNNs useful only for very short sequences. 

		\newpage
		\begin{figure}[h!]
    			\centering
			\subfloat[RNN]{{\includegraphics[width=.45\linewidth]{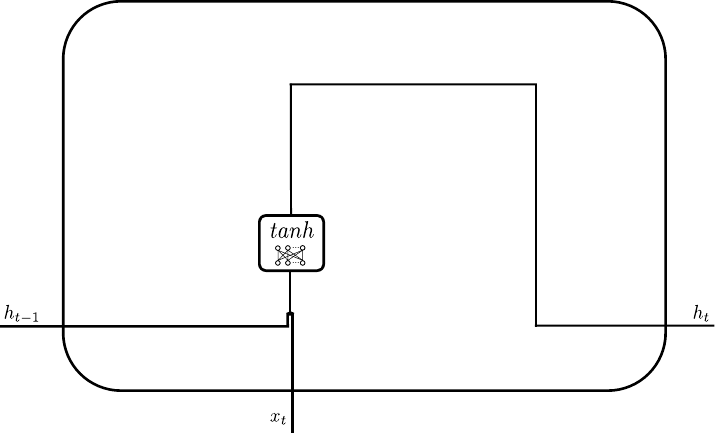} }}%
			\qquad
			\subfloat[LSTM]{{\includegraphics[width=.45\linewidth]{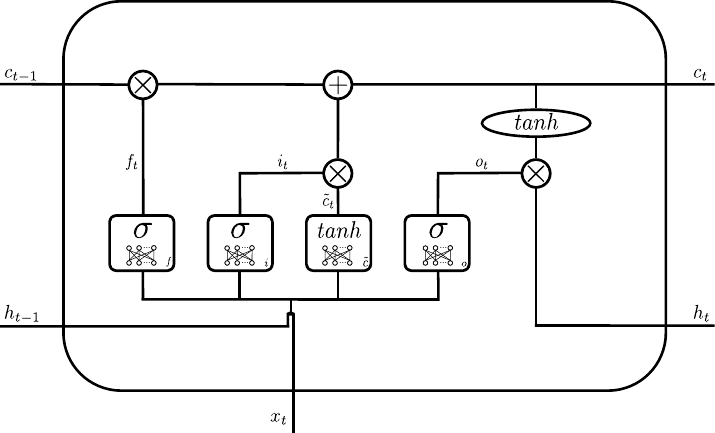} }}%
			\caption{Schematic representation of a classic RNN and an LSTM network}%
			\label{fig:RNN_LSTM}%
		\end{figure}

		To overcome (at least in part) the problem of short-term memory, the Long short-term memory (LSTM) \cite{LSTM} 
		architectures were introduced. Unlike a classic RNN network (Figure \ref{fig:RNN_LSTM}.a), the LSTM has a much 
		more complex single-cell structure (Figure \ref{fig:RNN_LSTM}.b), in which a single neural network is replaced by 
		four that interact with each other. However, the distinctive element of LSTM is the cell state $c$, which allows
		information to flow along the chain through simple linear operations. The addition or removal of information from 
		the state is regulated by three structures, called ``gates", each one with specific objectives. 

		The first, called ``forget gate," has the purpose of deciding the information that must be eliminated from the state.
		To reach this goal, the state is pointwise multiplied with:

		\begin{equation}
  			f_t = \sigma(W_f \cdot [h_{t-1},x_t] + b_f),
		\end{equation}
		
		\noindent
		that is obtained from a linear (actually affine because of the biases) block that combines the joint vector of the 
		input with the previous output ($[h_{t-1},x_t]$), followed by a standard logistic sigmoidal activation function
		($\sigma(x)={1 \over 1+e^{-x}}$). The forget gate makes  possible to delete (values close to zero), or to 
		maintain (values close to one), individual state vector components.		

		The second gate, called ``input gate", has instead the purpose of conditioning the addition of new information to 
		the state. This operation is obtained through pointwise multiplication between two vectors:
		
		\begin{eqnarray}
  			i_t &=& \sigma(W_i \cdot [h_{t-1},x_t] + b_i), \\
			\tilde c_t &=& \tanh(W_c \cdot [h_{t-1},x_t] + b_c)
		\end{eqnarray}

		\noindent
		the first (always obtained through a sigmoid activation) which decides the values to update, and the second 
		(obtained through a layer with $tanh$ activation) whose purpose is to create new candidates. Observe that the 
		function $tanh$ has also the purpose of regulating the flow of information, forcing the activations to remain in the
		interval $[-1; 1]$.

		\newpage
		Note that the cell status update depends only on the two gates just defined, and is in fact represented by the 
		following equation:

		\begin{equation}
  			c_t = f_t * c_{t-1} + i_t * \tilde c_t,
		\end{equation}
		
		\noindent
		Finally, there is the ``output gate", that controls the generation of the new output $h_t$ for this cell in relation to 
		the current input and previous visible state:

		\begin{eqnarray}
  			o_t &=& \sigma(W_o \cdot [h_{t-1},x_t] + b_o), \\
			h_t &=& o_t * tanh(c_t).
		\end{eqnarray}
		
		\noindent
		Also, in this case there is a sigmoid activation layer that determines which part of the state to send out, multiplying
		them by values between zero and one. Note that, again to limit output values, the $tanh$ function is applied to 
		each element of the state vector (this is a simple function without any neural layer) before it is multiplied by the 
		vector determined by the gate.

	\section{Implementation}
		Through the recurrent neural networks, with LSTM type architecture, the models used to achieve the intended
		objective are analysed and described below. Despite the relative simplicity of these architectures (from a general 
		point of view in the dynamic text processing), they prove extremely efficient in being able to catalogue the intent of 
		the answer; demonstrating how the decomposition of the general complex problem can also be tackled simply in 
		its individual parts.

		\subsection{Embedding}
			Obviously, and regardless of the type of neural network used, having to deal with text in natural language it 
			is essential to define the type of embedding used. In fact, unlike formal languages, which are completely 
			specified, natural language emerges from the simple need for communication, and is therefore the bearer of 
			a large number of ambiguities. To be able to at least try to understand it, it is therefore necessary to specify 
			a sort of ``semantic closeness" between the various terms, transforming the single words into vectors with 
			real values within an appropriate space \cite{Bengio2001}. The resulting embedding is therefore able to map 
			the single words into a numerical representation that ``preserves their meaning", making it, in a certain 
			sense, ``understandable" even to the computer. 

			Nowadays there are various ways to obtain this semantic space, generally known as Word Embeddings 
			techniques, each with its own peculiarities. For the prefixed purpose it was decided to use a pre-trained 
			embedding known as GloVe \cite{GloVe}, based on a vocabulary of 400,000 words mapped in a 
			300-dimensional space. 

		\subsection{Dataset}
			A heterogeneous dataset, consisting of questions both manually constructed and published by TREC and USC 
			\cite{Trec}, was used to train and test the various models. This dataset contains 5500 questions in English in 
			the training set and another 500 in the test set. Each question has a label that classifies the answer in a 
			two-level hierarchy. The highest level contains six main classes (abbreviation, entity, description, human, 
			location, numeric), each of which is in turn specialized in distinct subclasses, which together represent the 
			second level of the hierarchy (e.g. human/group, human/ind, location/city, location/country, etc). In total, 
			from the combination of all the categories and the sub-categories we get 50 different labels with which the
			answer to the supplied question is classified. An example extracted from the dataset is the following (label is 
			marked with bold and the question in italics):

			\vspace{1.5em}
			\textbf{HUMAN:ind} \quad \textit{Who was the 16th President of the United States?}
			\vspace{1.5em}
			
			\noindent
			As you can see, the label is composed of two parts separated by the symbol `:', representing the main class 
			and the sub class respectively. In this example, the main class indicates that the answer must communicate 
			a person or a group, while the sub class informs that we want to identify a specific individual (obviously 
			through his name).

			It should be noted that the representation provided by GloVe covers the totality of the words (9123 words) 
			present in the Dataset, without therefore the need for further work in this sense. However, once the data has
			been extracted from the dataset, a first manipulation is carried out, which consists in cleaning the strings
			from any special characters and punctuation symbols (excluding the question mark). Moreover, all the words
			contained in the strings are transformed into lowercase, thus avoid multiple codings for the same word.		

		\subsection{Models analysis and results}
			The search for the problem solution has been divided in two steps. It was in fact preferred to create a basic 
			model first, which aimed to classify only the main class, and then continue with a second model that also 
			used the secondary class.

			\subsubsection{First model}
\begin{figure}[h!]
					\centering
					\includegraphics[width=0.95\linewidth]{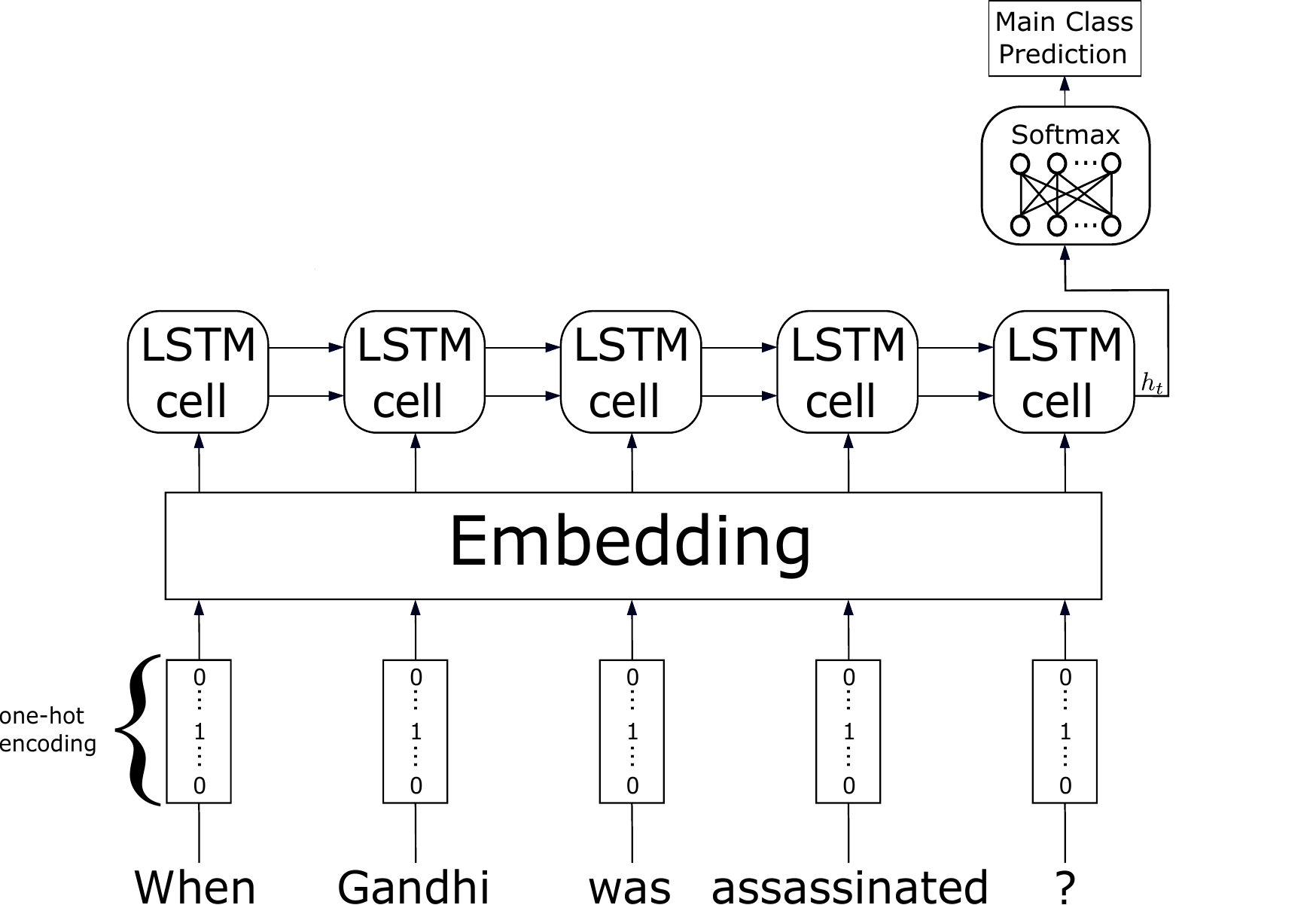}
					\caption{Representation of the first classification model}
					\label{fig:firstModel}
				\end{figure}

				To achieve the first objective, the strategy pursued was to map the entire sequence, entering the 
				network, into a vector of fixed dimensions, equal to the number of main categories. In other words, the 
				procedure foresees that every single word constituting the question (according to the temporal order 
				and after having been mapped through the level of embedding) is placed in input to the LSTM, whose 
				final state is mapped through a classical neural network with Softmax activation in one of the six main 
				prediction classes (Figure \ref{fig:firstModel}).

				The Table \ref{tab:firstModel} shows the results of the accuracy obtained, both on the training set and 
				on the test set, varying the size of the $h$ state of the LSTM.
				
				The supervised learning of the model was carried out using the Backpropagation through time (BPTT)
				\cite{BPTT} algorithm with the classical Categorical Cross-Entropy cost function:

				\begin{equation}
  					L(t,p) = -\sum_n \sum_j t_{n,j} log(p_{n,j})
				\end{equation}

				\noindent
				where the first summation is extended on the number of evaluated samples (examples) while the
				second on the number of classes owned; $t_{n,j}$ is the target of the example $n$ represented as 
				one-hot vector where only the $j$-th entry is $1$ and $p_{n,j}$ is the predicted probability that 
				example $n$ belong to class $j$.

				\begin{table}
					\begin{center}
						\setlength\tabcolsep{1.5em}
						\begin{tabular}{c|c|c}
							\hline
							h Dimension & Training set & Test set \rule{0pt}{12pt}\\[2pt]
							\hline
							25 & 99,26\% & 87,80\%\rule{0pt}{12pt}\\
							50 & 99,94\% & 89,80\%\\
							75 & 99,94\% & 90,60\%\\
							100 & 99,98\% & 90,20\%\\[2pt]
							\hline
						\end{tabular}
					\end{center}
					\caption{Accuracy relative to the size of the state for the first model}
					\label{tab:firstModel}
				\end{table}

			\subsubsection{Second model.}
				The good results obtained in the prediction of the main class, have further encouraged the progress in 
				the development of the model, maintaining  the first part unchanged. In this second part we have 
				included the subclass prediction, that having to represent a specialization of the main classes, it must be 
				influenced in some way by the prediction of the first.

				The basic idea was therefore simply to add a further element to the end of the question. 	This added 
				padding element does not carry any kind of information, but it is necessary only to evaluate the output 
				following the one corresponding to the last element of the question. The second last exit, linked to the 
				last element of the question, can be considered just like in the previous case, while the latter depends 
				only on the previous state (represented both by $h_{t-1}$ and $c_{t-1}$) since the embedding 
				representative of the padding element is previously set to a vector of all 	zeros and hence doesn't 
				contribute to computation of $h_t$ (look at Section \ref{sec:RNN-Intro}). 
		
\begin{figure}[h!]
					\centering
					\includegraphics[width=0.95\linewidth]{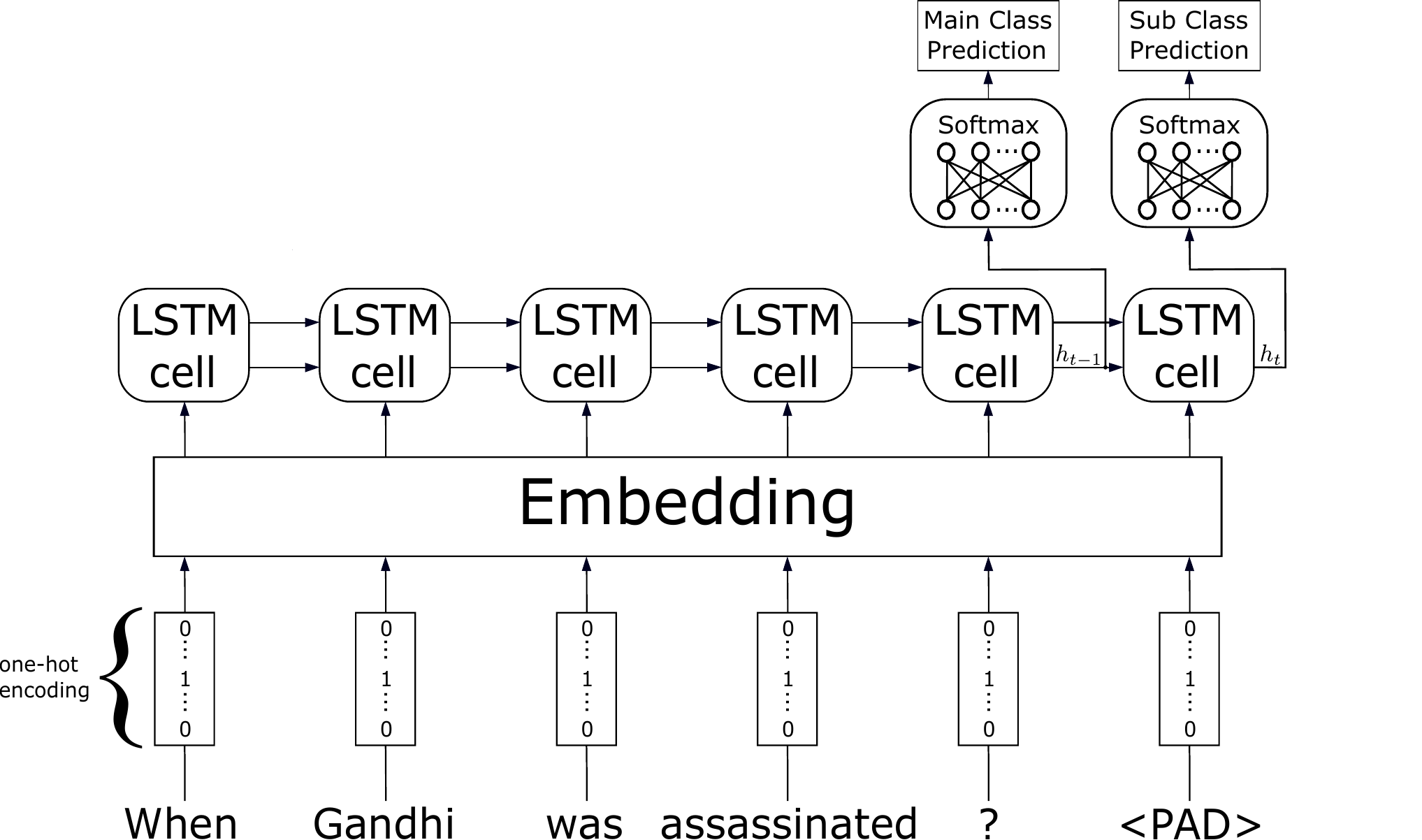}
					\caption{Representation of the double classification model}
					\label{fig:secondModel}
				\end{figure}

				By training the network, in a supervised manner, to associate the sub-category to this output, a 
				dependence will therefore be created on both the main classification and the question itself. The two 
				information coming out of the recursive part are subsequently discriminated by two distinct 
				fully-connected layers with Softmax, thus mapping the input sequence into two fixed size vectors 
				representing the main class and the associated sub class. The schematic representation of the second
				model just described can be observed in Figure \ref{fig:secondModel}.

				The BPTT algorithm is still used for the training phase of the model, but with a subtle difference 
				compared to the previous case: in fact, there will not be a single (Categorical Cross-Entropy) cost 
				function but two, since our goals have doubled. 

				Table \ref{tab:secondModel} shows the results obtained, with an accuracy of around $80\%$ for the 
				prediction of the sub-category of the samples belonging to the test set. Furthermore, as was to be 
				expected, there are almost identical performances on the main category prediction.

				\begin{table}
					\vspace{-1em}
					\begin{center}
						\setlength\tabcolsep{1.5em}
						\begin{tabular}{c|c|c|c|c}
							\cline{2-5}
							\multicolumn{1}{ c }{} & \multicolumn{2}{ c| }{Training set} & 
							\multicolumn{2}{ c }{Test set\rule{0pt}{12pt}} \\[2pt] \cline{1-5}
							\hline
							h Dimension & Main class & Sub class & Main class & Sub class \rule{0pt}{12pt}\\[2pt]
							\hline
							25 & 99,24\% & 96,86\% & 86,20\%  & 74,40\%\rule{0pt}{12pt}\\
							50 & 99,94\% & 99,83\% & 90,00\% & 80,00\%\\
							75 & 99,94\% & 99,72\% & 91,00\% & 78,60\%\\
							100 & 99,82\% & 99,67\% & 91,20\% & 82,20\%\\[2pt]
							\hline
						\end{tabular}
					\end{center}
					\caption{Accuracy relative to the size of the state for the second model}
					\label{tab:secondModel}
					\vspace{-2em}
				\end{table}

\begin{figure}[h!]
					\begin{minipage}[b]{.48\linewidth}
						\centering
						\centerline{\includegraphics[width=1.0\linewidth]{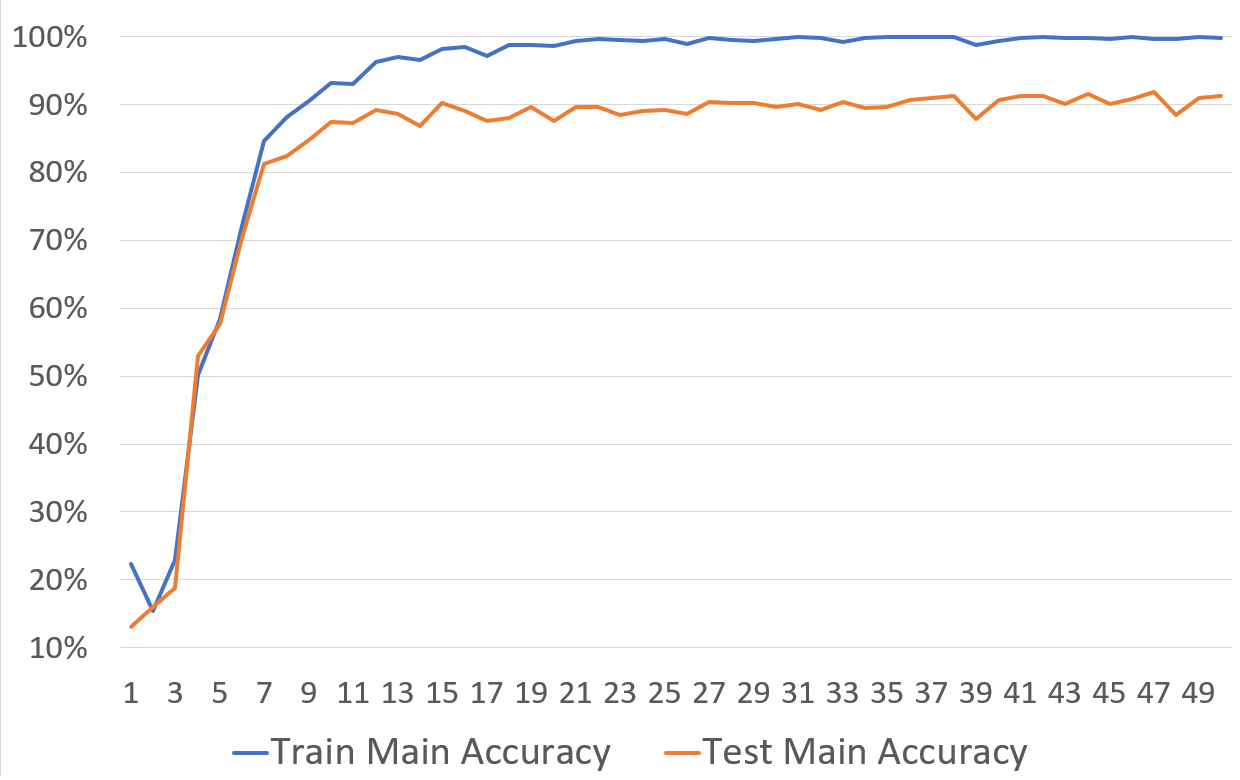}}
						\centerline{(a)}\medskip
					\end{minipage}
					\hfill
					\begin{minipage}[b]{0.48\linewidth}
						\centering
						\centerline{\includegraphics[width=1.0\linewidth]{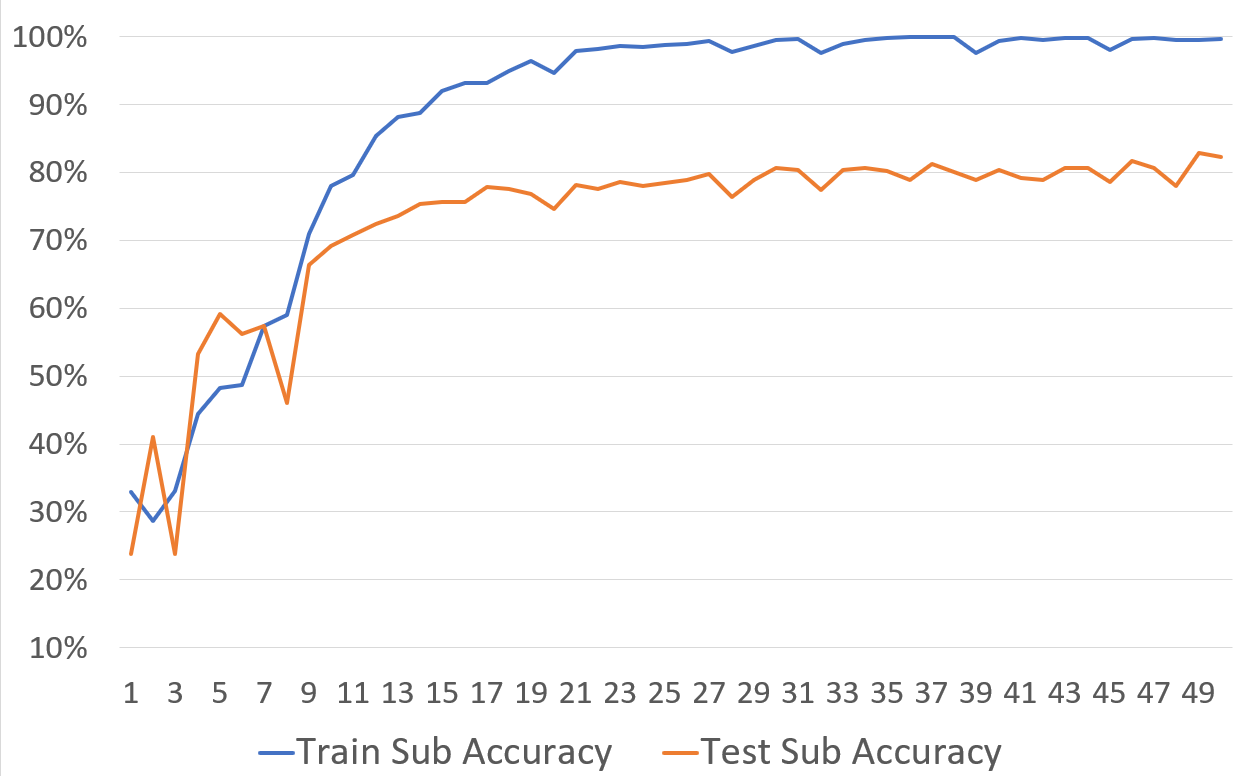}}
						\centerline{(b)}\medskip
					\end{minipage}
					\vspace{-1em}
					\caption{Accuracy trend for the Main (a) and Sub (b) classes at different epochs.}
					\label{fig:4overfitting}
					\vspace{-2em}
				\end{figure}

				It should be noted that, despite the excellent performance of the model on the training set, it does not 
				seem to have gone into overfitting. This affirmation can be confirmed by observing through the 
				accuracy trend for the model with $H=100$ (Figure \ref{fig:4overfitting}) both on the training and on
				the test set at different epochs, which shows how the greater accuracy on the training set does not 
				become pejorative for the test set. In fact, both trends seem to stabilize around a regime value, with 
				slight fluctuations that do not seem to affect the generalization characteristics of the network.

		\subsection{Prototype responder}
\begin{figure}[h!]
				\centering
				\includegraphics[width=0.9\linewidth]{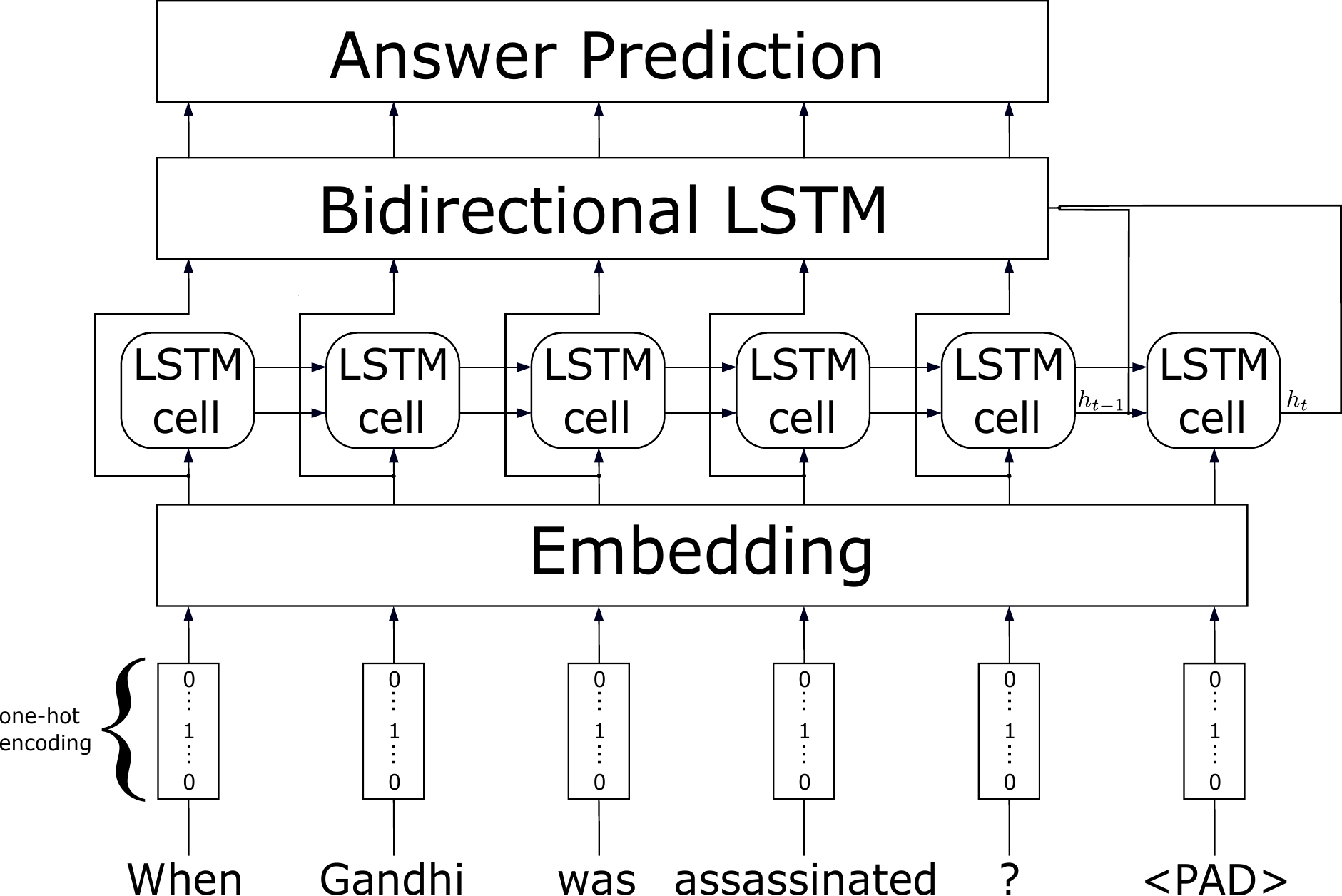}
				\caption{Representation of the responder prototype}
				\label{fig:responder}
				\vspace{-1em}
			\end{figure}

			Finally, in order to test (at least briefly) the importance of what was achieved in the classification of intents
			on the ultimate purpose of creating a responder, it was decided to create a prototype that could exploit the 
			categorization obtained to generate a response. This prototype (Figure \ref{fig:responder}) uses a 
			bidirectional LSTM (BLSTM) network \cite{BRRN} to review the inbound application, so that it can generate 
			an answer only after acquiring the entire question. 

			The status of this BLSTM network is conditioned by the prediction returned by the previous model. The two 
			vectors representing the main and the sub class are in fact linked together, forming a single vector that
			will represent the initial state of the BLSTM network. The exits of the network, which constitute the words 
			forming the answer, are thus generated by the analysis of both contexts (future and past), and strongly 
			influenced by the categorization provided.

			\begin{table}
				\vspace{-1em}
				\begin{center}
					\setlength\tabcolsep{1.5em}
					\begin{tabular}{r|l}
						\hline
						\multicolumn{1}{ c }{Question} & \multicolumn{1}{ c }{Answer} \rule{0pt}{12pt}\\[2pt]
						\hline
						How many people speak French? & 13\rule{0pt}{12pt}\\
						What day is today? & the first day may nights and the \\
						Who will win the war? & north \\
						Who is Italian first minister? & francisco vasquez \\
						When World War II ended? & march \\
						When Gandhi was assassinated? & 1976\\[2pt]
						\hline
					\end{tabular}
				\end{center}
				\caption{Examples of answers for questions not belonging to the training set}
				\label{tab:QA}
				\vspace{-2em}
			\end{table}

			The supervised training of the network was performed on a set of 500 question-answer samples, and in the 
			Table \ref{tab:QA} are shown some of the network outputs relating to questions not present in the training 
			set. It should be noted that the purpose of this prototype is not to provide a correct answer (which is quite 
			impossible given the limited dataset and since no knowledge of the answer not relating to it is never provided 
			to the network) but to show how the simple excellent categorization of the intent allows, already alone, to get 
			consistent answers to the context of the question.

	\section{Conclusion}	
		The results obtained from the models presented show very high accuracy values, both for the training set and for 
		the test set. Networks of this type are actually very effective in this type of classification, perhaps because of their 
		simplicity. The example of the responder prototype then confirms how the choice to classify the questions based 
		on the intent of the answer is extremely effective in limiting and contextualizing the outgoing answer. 
		Consequently, this approach can be of help for the development of more complex systems, being already able to 
		compensate for the deficiencies of ``traditional" systems that can benefit from the classification of intent provided.
	
	\bibliography{paperBib}{}
	\bibliographystyle{ieeetr}

\end{document}